\pdfoutput=1

\documentclass[11pt]{article}

\usepackage[final]{acl}

\usepackage{times}
\usepackage{latexsym}
\usepackage{subfiles}

\usepackage[T1]{fontenc}

\usepackage[utf8]{inputenc}

\usepackage{microtype}

\usepackage{inconsolata}

\usepackage{graphicx}
\usepackage{tcolorbox}
\usepackage{cleveref}
\usepackage{acronym}
\acrodef{lsc}[LSC]{Lexical Semantic Change}
\acrodef{nlp}[NLP]{Natural Language Processing}
\acrodef{lm}[LM]{Language Model}
\acrodef{llm}[LLM]{Large Language Model}
\acrodef{kg}[KG]{Knowledge Graph}
\acrodef{cot}[CoT]{Chain-of-Thougth}
\acrodef{wsd}[WSD]{Word Sense Disambiguation}

\title{Semantic Change Characterization with LLMs using Rhetorics}

\author{
 \textbf{Jader Martins Camboim de S\'a\textsuperscript{1,2}},
 \textbf{C\'edric Pruski\textsuperscript{2}},
 \textbf{Marcos Da Silveira\textsuperscript{2}}\\
\textsuperscript{1}FSTM - University of Luxembourg\\
  2 place de l’Université, L-4365, Esch-sur-Alzette, Luxembourg,\\
 \textsuperscript{2}Luxembourg Institute of Science and Technology\\
        5 avenue des Hauts-Fourneaux, L-4362, Esch-sur-Alzette, Luxembourg,
\\
 \small{
   \textbf{Correspondence:} \href{mailto:email@domain}{first.second@list.lu}
 }
}

\begin{document}
\maketitle
\begin{abstract}
Languages continually evolve in response to societal events, resulting in new terms and shifts in meanings. These changes have significant implications for computer applications, including automatic translation and chatbots, making it essential to characterize them accurately. The recent development of \acp{llm} has notably advanced natural language understanding, particularly in sense inference and reasoning. In this paper, we investigate the potential of \acp{llm} in characterizing three types of semantic change: dimension, relation, and orientation. We achieve this by combining \acp{llm}' Chain-of-Thought with rhetorical devices and conducting an experimental assessment of our approach using newly created datasets. Our results highlight the effectiveness of \acp{llm} in capturing and analyzing semantic changes, providing valuable insights to improve computational linguistic applications. 
\end{abstract}

\section{Introduction}
\label{sec:intro}
Language, a tool humans acquired throughout evolution, remains a subject of fascination and inquiry across diverse disciplines, including neuroscience, psychology, philosophy \cite{pinker2003language}, and computational linguistics. Despite this interdisciplinary interest, our understanding of language is often superficial, with much to uncover regarding its intricacies \cite{allan2013oxford, Pinker1990NaturalLA}. Among the many elements that shape language, a central aspect in understanding dynamics in language development is how the semantics of words change \cite{allan2013oxford, Pinker1990NaturalLA}. This evolution is particularly intriguing in computational linguistics, as it impacts applications such as automatic translation and chatbots \cite{camboim-de-sa-etal-2024-socio}. While humans can rapidly adapt to changes using a lot of contextual information and cognitive processes to grasp the senses of a sentence or a word, it is complex to provide enough cultural knowledge and nuances to machines. Consequently, machines lack the tools to adapt to these variations and to perform effective communication \cite{Tahmasebi2018SurveyOC}. Therefore, in many modern \ac{nlp} systems, we observe the impacts of semantic change on end users, especially when the task requires deep contextual dependency \cite{camboim-de-sa-etal-2024-socio}.

In the context where historical or domain-specific knowledge of meaning is crucial, the \ac{lsc} field emerged to gain deep understanding of and detect these changes \cite{Tahmasebi2018SurveyOC}. While a significant body of work explores which words changed in different moments or domains, there is still a need for further comprehension regarding the types and implications of semantic changes in these systems \cite{Hengchen2021ChallengesFC}. For instance, in sentiment analysis, being aware that the term `sick' has acquired a positive connotation could significantly alter the interpretation of a sentence. 

Theories to comprehend semantic change exist. One prominent typology, proposed by \cite{traugott2017semantic}, categorizes change into broadening/narrowing (a word gains or loses senses), amelioration/pejoration (a word is perceived more positively/negatively), metaphorization and metonymization (the word is used as a metaphor or metonymy respectively). We illustrate these types of change in \Cref{tab:examples}.

\begin{table}[ht]
    \centering
    \resizebox{\columnwidth}{!}{\begin{tabular}{lr}\hline
        A child in \textbf{dirty} overalls. &\\
        He used a \textbf{dirty} trick to win the competition.& pejoration\\\hline
        No other style of \textbf{hat} was acceptable with an evening dress.&\\
        He took off his politician's \textbf{hat} and talked frankly.&metaphorization\\\hline
        The diamond is currently set in the \textbf{crown} of the Queen.&\\
        The colonies revolted against the \textbf{crown}.& metonymization\\\hline
    \end{tabular}}
    \caption{Examples illustrating the characterization of types of change.}
    \label{tab:examples}
\end{table}


Previous works have only partially covered the typology of semantic change, typically focusing on a few types \cite{Sa2024SurveyIC}. However, recent advancements in Language Models (\acp{llm}) have showcased capabilities in executing complex linguistic tasks such as inference, association, understanding, and common sense reasoning \cite{Achiam2023GPT4TR} via the Chain-of-Thought (\ac{cot}) technique \cite{Wei2022ChainOT}. These reasoning abilities in \acp{llm} mimic human-like processes for establishing connections and relations in natural language \cite{Dasgupta2022LanguageMS}.

Rhetorical devices, known for their role in building persuasive arguments and reasoning, utilize cognitive processes to improve argumentation and communication efficiency \cite{lakoff2008metaphors}. These devices facilitate concise communication and also exhibit characteristics related to semantic evolution over time. Hence, they have been extensively employed by human evaluators to compare senses \cite{kearns2006lexical, Steen2007MIPAM}. In this process, evaluators use historical and cultural knowledge to explain variations in semantic change through rhetorical argumentation.

Moreover, recent studies have found that \acp{llm} encode extensive cultural knowledge, including relationships, associations, and events \cite{petroni-etal-2019-language}, making them suitable for automating the characterization of semantic change. Building upon this insight, we propose leveraging rhetoric in natural language to characterize semantic change within the proposed typology by exploring \acp{llm} ``thought'' processes to mimic human cognitive reasoning. Following the outlined typology, our methodology aims to instruct \ac{llm} to utilize rhetorical devices and characterize change within a comparative framework. Our contributions are:

\begin{itemize}
    \item A new approach to semantic change characterization exploring ``reasoning'' and rhetoric capabilities of \acp{llm}.
    \item The proposal of 3 new public datasets for evaluation of semantic change characterization: dimension, orientation, and relation.
\end{itemize}

The paper is structured as follows: \Cref{sec:related} presents related work of the field semantic change characterization using \acp{llm}. \Cref{sec:method} details our methodology for prompting models for semantic change identification and characterization. \Cref{sec:expe} introduces the experimental settings, including datasets and results. \Cref{sec:discussion} discusses insight from the method. Finally, \Cref{sec:conclusion} contains concluding remarks and outlines future work.

\section{Related Work}
\label{sec:related}
Most of the papers in semantic change address the problem of identification, i.e., detecting if the meaning of a word changed without inferring what type of change occurred. In the context of \acp{lm} some authors explore these capabilities to track semantic change identification as a sequence-to-sequence problem \cite{lyu-etal-2022-mllabs, giulianelli-etal-2023-interpretable}, by first prompting the model to disambiguate the word in context and then generating a contextualized word representation.

In challenges for semantic change identification for Russian and Spanish \cite{Pivovarova2021RuShiftEvalAS, ZamoraReina2022blackLSCDiscoveryST}, the best performing methodologies were large cross-language models fine-tuned in \ac{wsd} for English data to then fit a linear regression over the contextualized embeddings to identify semantic change. Later, \acp{llm} were employed for this task \cite{Wang2023LargeLM, Periti2024AnalyzingSC}, but using only few-shot prompt or fine-tuning to the task.

Previous works on semantic change characterization relies on extracting word representations for each corpus, and later compare them to capture possible differences in usage \cite{Sa2024SurveyIC}. In the context of broadening/narrowing \citet{Bochkarev2022Neural} utilize a neural network to determine if a word is employed as a named entity. This approach creates a temporal perspective of a word's usage and allows them to compare occurrences of a word to see whether it has gained new usage in the corpus. For metaphorization \citet{maudslay-teufel-2022-metaphorical} fine-tuned with supervision a BERT model to classify contextualized words into metaphor and then analyzed different corpus. Finally, in the amelioration/pejoration context \citet{Fonteyn2021AdjustingSA} measures polarity in the term `to death' by calculating the distance between the word vectors `good' and `bad'.

Compared to previous works on semantic change, this is the first study to use \ac{cot} for this task, with our approach being deeply motivated by linguistic literature. In terms of semantic change characterization, this is the first work that generalizes across all types of change \cite{Sa2024SurveyIC}, has no dependency on training data, and can be used for every type of relation e.g., metaphor and metonymy.

\section{Methodology}
\label{sec:method}

\subsection{Background}
\label{sec:background}
In this paper, we propose a method for automating the characterization of semantic change across different corpora. To this end, we rely on the following set of predominant typologies defined in the literature \cite{traugott2017semantic, juvonen2016lexical}:
\begin{itemize}
    \item \textbf{Broadening:} gaining a new meaning related or not to the previous meaning, such that a word represents more concepts, e.g., `cloud,' a computing infrastructure.
    \item \textbf{Narrowing:} restriction of meaning occurs when a symbol represents fewer concepts than previously, e.g., `gay' which historically meant festive or happy, is now predominantly used to refer to homosexuality.
    \item \textbf{Amelioration:} a word gains a more positive sense to the previous sense, \textit{nice}, `foolish, innocent' changed to `pleasant.'
    \item \textbf{Pejoration:} the word is used with a worse connotation to the previous usage, \textit{stincan}, `smell (sweet or bad)' changed to \textit{stink}.
    \item \textbf{Metonymization:} association between terms, e.g., \textit{board} `table', changed to ``people sitting around a table, governing body.''
    \item \textbf{Metaphorization:} conceptualizing one thing in terms of another, e.g., `head of the company' the word `head' conceptualizes ``command or control.''
\end{itemize}

\begin{figure}[ht]
    \centering
    \includegraphics[width=0.8\columnwidth]{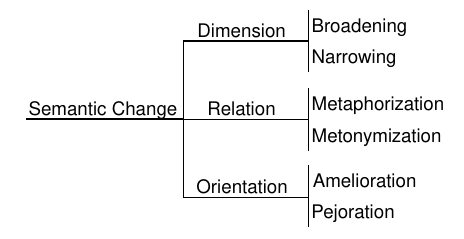}
    \caption{Taxonomy for the poles of Lexical Semantic Change \cite{traugott2017semantic, juvonen2016lexical}.}
    \label{fig:taxonomy}
\end{figure}

We used the same classification of typologies as presented by \citet{Sa2024SurveyIC}, where the typologies can be regrouped into three poles, namely Dimension, Relation, and Orientation (see \Cref{fig:taxonomy}).

In the \textbf{dimension} pole, we compute the ``number of senses'' a word can have. This pole is self-complementary as increasing represents broadening, and decreasing represents a narrowing of senses. After identifying the number of senses, we can compare the differences between corpora.

Metaphorical and metonymical changes are classified under the \textbf{relation} category, as these changes enhance the connection between one sense of a word and its other senses. In this framework, a word's meaning relies on the link established through either conceptual (abstract relation) or material (physical association) similarity between concepts. We identify which senses are used figuratively in relation to other senses of the same word.

The \textbf{orientation} pole regroups the process of amelioration or pejoration of a meaning. In this pole, words are analyzed according to the contextual sentiment captured from each corpus, and then we analyze how the sentiment changes over corpora. In this study, we explore only positive, negative, and neutral sentiment values for words.

\subsection{Rhetorical Arguments as a Pragmatic Tool}
\acp{llm} have exhibited significant progress in natural language comprehension. This includes reasoning by analogy \cite{Webb2022EmergentAR}, understanding metaphors \cite{liu-etal-2022-testing}, argumentation \cite{Chen2023ExploringTP}, and acquiring cultural knowledge \cite{petroni-etal-2019-language}. Additionally, instructing an \ac{llm} to generate a rationale, which is a natural language explanation for its reasoning process, before providing an answer has been shown to improve performance on many \ac{nlp} tasks that require logical reasoning \cite{Wei2022ChainOT, kavumba-etal-2023-prompting}. This rationale generation step is believed to inject more information retrieved from the \ac{llm}'s internal knowledge store into the prompt. This enriched prompt allows the \ac{llm} to consider a broader range of knowledge during the final decision-making process \cite{Dasgupta2022LanguageMS}.

In this paper, we rely on its stored cultural knowledge to improve the context for the task and work as an annotator in the framework proposed in the previous subsection. The problem of characterizing \ac{lsc}, identifying different senses, figurative usage, and feelings, relies on building cognitive relations between other senses that depend on human perception and culture \cite{lakoff2008metaphors}. As language is the best tool to explain language \cite{pinker2003language}, in our approach, we try to mimic this cognitive process \cite{huang-chang-2023-towards}, using the human knowledge contained in \acp{llm} and rationales as a means to produce the same associations human perform \cite{Dasgupta2022LanguageMS, Strachan2024TestingTO}.

Tracking word senses and comparing them is a complex task \cite{kilgarriff1997don}. To address this problem, we approach it as a ``comparative semantics'' problem, i.e., instead of extracting the meaning as a final objective, we rely on the idea of relatedness and likeness of meaning to compute \ac{lsc}. Similar to the work of \citet{schlechtweg-etal-2024-durel}, we compare word occurrences for the characterization problem. However, in DUREL the annotator is prompted with two sentences that share a particular word and the annotator has to classify the level of similarity between contextual senses. In our approach, we use an \ac{llm} instead of humans to annotate the relatedness between words, and we reduce it to just `identical' and `different' classes.

We perform the characterization starting with a cognitive semasiological comparative analysis of the word meaning \cite{kilgarriff1997don}, following the Cambridge setting \cite{Tang2013SemanticCC}. We first provide a context where the word sense could be inferred (the Gracian approach \cite{agirre2007word}), to then decide via reasoning if the deduced senses are different to a class of change \cite{blank2003words}.

This step is done by a \ac{llm} that acts like a judge/annotator using a special type of \ac{cot} \cite{Wei2022ChainOT} with detailed step-by-step reasoning \cite{Mitra2023Orca2T} to elicit models ability of word sense induction and comparative semantics. Our approach exploits rhetorical techniques to produce `cognitive-appealing' arguments on how the senses are different.

For the first pole, \textbf{dimension}, we created a prompt requesting a word sense differentiation, where the prompt ask if a word is used in an identical or different sense. To perform the sense differentiation, we instruct the model to use \textbf{zeugma} \cite{kearns2006lexical} as a cognitive approach to identify identical senses. If it can produce a consistent zeugma, the senses are identical. Otherwise, it should assume the words are different.

Zeugma is a rhetorical device where a single word, typically a verb or an adjective, governs or modifies two or more words in a sentence. This device creates a clever or unexpected relationship between different sentence parts. Zeugma often results in a play on words contributing to the overall impact of the expression, and it adds a layer of complexity or humor to the language used in a sentence and allows us to explore the sense usage difference \cite{kearns2006lexical}. For example:

\begin{quote}
    (1) "He lost his keys."\\
    (2) "He lost his temper."\\
    (3) "He lost his keys and his temper." (?)
\end{quote}

In sentence (3), the word ``lost'' is used to combine both sentences in a related sense to describe both (1) losing physical objects (keys) and (2) losing emotional control (temper). This zeugma creates a figurative and compact expression that links two different related meanings of the word `lost' in a single sentence, creating a bad pun \cite{kearns2006lexical}. This bad pun comes because the second usage of 'lost' does not preserve the same sense as the first usage, indicating a difference in the meaning. In \Cref{fig:dim}, we present part of the prompt employing zeugma for the dimension dataset. In \Cref{sec:prompt}, we share the complete prompt for the experiments. 

\begin{figure}[ht]
    \centering
    \begin{tcolorbox}[title=Sense Differentiation, fontupper=\small]
    [...]. Follow these steps to complete the task:
    \begin{itemize}
    \item Step 1. Describe the meaning of the word in the first sentence.
    \item Step 2. Describe the meaning of the word in the second sentence.
    \item Step 3. Write a sentence that joins both using zeugma and the same shared word while preserving the same sense. If the construction makes a bad pun, the words have a different sense.
    \item Step 4. Based on the previous reasoning, give your final answer: 'identical' or 'different.'
    \end{itemize}
    \tcblower
    \textcolor{gray}{[Few-shot examples.]}
    \end{tcolorbox}
    \caption{Prompt for sense differentiation in the dimension dataset.}
    \label{fig:dim}
\end{figure}

This work proposes a computational instruction for figurative language analysis. The instruction is based on a simplified version of the Metaphor Identification Procedure (MIP) \cite{Steen2007MIPAM}. To distinguish between metaphor and metonymy, the model is tasked with building a relation between the concepts. This relation can be either abstract, suggesting a metaphorical mapping between domains (evoking tropes), or material, if a physical association exists between the concepts.

\begin{quote}
    (1) "The main objective of this forthcoming decision will be preparation for the winter."\\
    (2) "Winter can cause many disruptions for public transport."\\
    (3) "The word 'winter' in the second sentence is associated with its problems, such as snow, making it a metonymy."
\end{quote}

The provided examples showcase how \textbf{simile} act as a parsing mechanism for the AI model. By leveraging similes, the model can reframe figurative language based on the similarity or association it expresses between concepts. In essence, a simile acts as a rhetorical device that explicitly compares two entities to enhance explanation and detail the nature of that comparison.  By deciphering the figurative meaning within context, we aim to guide the model towards extracting more information about the underlying semantic relationship. This, in turn, allows the model to make a more accurate prediction regarding the type of relation – whether it's a metaphor or metonymy. In \Cref{fig:rel}, we illustrate the prompt for obtaining the figurative association between word usages.

\begin{figure}[ht]
    \centering
    \begin{tcolorbox}[title=Sense Figurativeness,fontupper=\small]
    [...]. Follow these steps to complete the task:
    \begin{itemize}
    \item Step 1. Describe the meaning of the word in the first sentence.
    \item Step 2. Describe the meaning of the word in the second sentence.
    \item Step 3. Compare the usage, determining if the second is related as a metaphor (where the word is used in a similar but non-literal sense), as a metonymy (where the word represents something closely related to or associated with it), or unrelated, used with a different sense.
    \item Step 4. Based on the third step, write the final answer, 'metaphor', 'metonymy', or 'unrelated.'
    \end{itemize}
    \tcblower
    \textcolor{gray}{[Few-shot examples.]}
    \end{tcolorbox}
    \caption{Prompt for figurative sense in the relation dataset.}
    \label{fig:rel}
\end{figure}

For \textbf{orientation} pole, the current state-of-the-art sense-level sentiment analysis requires first a \ac{wsd} step, then a sentiment analysis step \cite{Zhang2023NeuroSymbolicSA}. Similar to the previous prompts and following the best practice, we instruct the rationale to perform a textual sense disambiguation and then differentiate the orientations between these senses \cite{wiebe-mihalcea-2006-word}. To differentiate orientation, we use \textbf{antanagoge} to compare senses' positiveness (or negativeness) and to enrich contextual information on how these senses can be perceived in the training data.

Antanagoge is a rhetorical device that involves responding to an accusation or negative point with a counter-argument or positive point. It is used to mitigate the impact of something negative by placing it alongside something positive.
We use the common sentence ``I'd rather X than Y'' as a few-shot demonstration to instruct \ac{llm} to get the most probable contextual ordering. In the example below exemplify the usage of antanagoge.

\begin{quote}
    (1) "A terrific presentation."\\
    (2) "A terrific storm."\\
    (3) "I'd rather have an terrific presentation than an terrific storm."
\end{quote}

\begin{figure}[ht]
    \centering
    \begin{tcolorbox}[title=Sense Orientation,fontupper=\small]
    [...]. Follow this instructions to execute the task:
    \begin{itemize}
    \item Step 1. Describe the meaning of the word in the first sentence.
    \item Step 2. Describe the meaning of the word in the second sentence.
    \item Step 3. Leverage the rhetorical strategy of antanagoge, contrasting a negative with a positive, to weigh why one meaning might be more favorable than the other, or if they stand neutral.
    \item Step 4. Based on the third step, write the final answer 'negative', 'positive', or 'neutral.'
    \end{itemize}
    \tcblower
    \textcolor{gray}{[Few-shot examples.]}
    \end{tcolorbox}
    \caption{Prompt for sense orientation in the orientation dataset.}
    \label{fig:ori}
\end{figure}

Further optimization can be done as in \cite{schlechtweg-etal-2024-durel}. A graph of occurrences can be build and clusterized to extract senses over time. These clusters allows annotating semantic changes for each sense over time.

\section{Experiments}
\label{sec:expe}

In this section, we introduce the dataset produced to evaluate our annotation method and measure the quality of our annotations for \ac{lsc}.

\subsection{Dataset for Lexical Semantic Change Characterization}
\label{sec:dataset}
From a semasiological perspective, words' meaning could be inferred from the context, for example, ``He \textit{targeted} me, after I didn't agree with his proposal.'' or ``The \textit{mustache} guy, is coming today?''. We can deduce the meaning of a word based on context and/or knowledge of the original meaning.

Lexical and Semantic Change (LSC) reflects how word meaning evolves. New senses emerge when a word is used in a novel, non-standard way. Over time, if this usage becomes widespread enough, it transitions from creative expression to a conventional meaning.

The ideal approach to LSC detection should mimic this human capability.  This means employing unsupervised learning techniques, where the system infers the evolving sense of a word solely based on its prior exposure to the language and the contextual information within the data. In essence, the system learns to identify semantic shifts without the need for pre-labeled data \cite{schlechtweg-etal-2020-semeval}.

To evaluate our framework, we produce an \ac{lsc} Characterization dataset following the Cambridge setting described in \cite{Tang2013SemanticCC}. The dataset is composed of pairs of sentences sharing the same word (see the example with the word 'lost' from equation (1) and (2)). The first sentence expresses one possible usage (e.g., original usage), while the second sentence express a different usage. The task is to infer how the word's meaning in each context and compare them. We create three new datasets, one for each pole of change, where the instance pairs present the type of change the \ac{llm} should characterize in the pipeline.

In the \textbf{dimension} dataset, we curated the WiC data \cite{pilehvar-camacho-collados-2019-wic} for getting a fraction of reliable and high-quality examples. The original dataset only classified the word's meaning as related or unrelated. We adapted it according to the DUREL format. For this, the word's meanings are identical if the same meaning is observed when we merge the sentences (see the 'lost' example, equation (3)), i.e., a zeugma can be performed between the two sentences, relabeling the sentences to this.

We define words as related if they have a direct relation (metaphor or metonymy) for their usage, if the relation is not direct, we define this senses as unrelated. For example, if `head' is used figuratively in both usages, the `head' to represent the leader of a company and the `head' to represent the mind, we define them as unrelated as there is no direct figurative usage between them. Finally, if not one of the cases above, we set them as unrelated or keep the original annotation.

For the \textbf{relation} data, we collected examples from the metaphor detection dataset \cite{choi-etal-2021-melbert} to get literal and metaphorical usages and also examples from the literature to increase the evaluation dataset size, the sentences were manually collected using online dictionaries like Linguee\footnote{\url{www.linguee.com}} and Merriam-Webster\footnote{\url{www.merriam-webster.com}} and verified by 3 human annotators. To collect metonymies, we similarly used examples in literature \cite{lakoff2008metaphors} and retrieved sentences from online sources.

The \textbf{orientation} data we created by getting sense pairs for the same word where we had the greatest variance from these pairs by analyzing SentiWordNet \cite{baccianella-etal-2010-sentiwordnet}. The sentence pairs were obtained from SemCor \cite{raganato-etal-2017-word} and WordNet \cite{Miller1995WordNetAL} depending on how easy it is to infer the sense given the sentence. Additionally, we transform the sentences so that the sentiment of a word cannot be trivially detected from the whole sentence, so the detector needs to comprehend the word-level sentiment. In some cases, we modified the sentence to be negative while the word meaning is positive.

\begin{table}[ht]
    \centering
    \resizebox{\columnwidth}{!}{\begin{tabular}{|c|c|c|}\hline
        Task & labels  & Total \\\hline
        Dimension & Identical , Different & 260 \\\hline
        Relation & Metaphor, Metonymy, Unrelated & 331 \\\hline
        Orientation & Positive, Negative , Neutral & 262 \\\hline
    \end{tabular}}
    \caption{General view of the three datasets created for Semantic Change Characterization.}
    \label{tab:dataset_stats}
\end{table}

In \Cref{tab:dataset_stats} we describe the number of instance pairs for each dataset we produced. 

\subsection{Experimental results}
\label{sec:results}
We compare our approach with two baselines to evaluate how good \acp{llm} and rhetoric devices are for characterizing semantic change. The prompt is based on \ac{cot} and a few-shot prompt with no \ac{cot}, where all prompts are provided with the same 3-shot examples with the correct label and also the dictionary sense. We took a special care to the difference in prompt be only the method and not inserting `hack' phrases\footnote{e.g. "Please think carefully", "Take a deep breath", "This is important for my career."} to improve model performance.

We selected \texttt{LLaMA-3} and \texttt{Phi-3} as the current state-of-the-art \acp{llm} for the instruction prompt. We sampled the models 5 times for each method with temperature $\tau=0.7$, using the guidance\footnote{\url{https://github.com/guidance-ai/guidance/}} library to control the generated layout. We report the mean and standard deviation of the accuracy. 

In \Cref{tab:dim}, we present the results for the dimension dataset. We can observe that the rhetoric method meaningfully improves the accuracy of \texttt{Phi-3} and \texttt{LLaMA-3-70b} over the baselines, while for \texttt{LLaMA-3-8b}, the best method is few-shot prompt. While the data used for instruction tuning \texttt{LLaMA-3} is not publicly released, we believe it was fine-tuned in WiC data \cite{pilehvar-camacho-collados-2019-wic}, which could explain the improved accuracy.

\begin{table}[ht]
\centering
\resizebox{\columnwidth}{!}{\begin{tabular}{|l|c|c|c|}
\hline
Method           & \texttt{LLaMA-3-8b}&\texttt{LLaMA-3-70b} & \texttt{Phi-3} \\ \hline
Few-Shot         & \textbf{.75$\pm$.00}& .76$\pm$.00& .60$\pm$.00\\ \hline
CoT              & .59$\pm$.02& .75$\pm$.02& .62$\pm$.01\\ \hline
Rhetoric (ours) & .68$\pm$.03& \textbf{.78$\pm$.01}& \textbf{.71$\pm$.02}\\ \hline
\end{tabular}}
\caption{Accuracy metrics for the Dimension dataset.}
\label{tab:dim}
\end{table}

\Cref{tab:rel} shows our results for the relation dataset. For both models, the rhetoric method improved significantly over the baselines. With \texttt{LLaMA-3-70b} getting the overall best capabilities for recognizing figurative usage.

\begin{table}[ht]
\centering
\resizebox{\columnwidth}{!}{\begin{tabular}{|l|c|c|c|}
\hline
Method           & \texttt{LLaMA-3-8b}&\texttt{LLaMA-3-70b} & \texttt{Phi-3} \\ \hline
Few-Shot         & .38$\pm$.00& .52$\pm$.00& .49$\pm$.00\\ \hline
CoT              & .45$\pm$.01& .53$\pm$.01& .51$\pm$.03\\ \hline
Rhetoric (ours) & \textbf{.52$\pm$.02}& \textbf{.59$\pm$.02}& \textbf{.56$\pm$.01}\\ \hline
\end{tabular}}
\caption{Accuracy metrics for the Relation dataset.}
\label{tab:rel}
\end{table}

In \Cref{tab:ori}, the accuracy is tied between few-shot and rhetoric for \texttt{LLaMA-3-8b}, while for \texttt{Phi-3} is 10 points better. This task is particularly more difficult than the rest for \acp{llm}.

\begin{table}[ht]
\centering
\resizebox{\columnwidth}{!}{\begin{tabular}{|l|c|c|c|}
\hline
Method           & \texttt{LLaMA-3-8b} &\texttt{LLaMA-3-70b}& \texttt{Phi-3} \\ \hline
Few-Shot         & \textbf{.37$\pm$.00}& .36$\pm$.00& .30$\pm$.00\\ \hline
CoT              & .34$\pm$.03& .35$\pm$.00& .30$\pm$.02\\ \hline
Rhetoric (ours) & \textbf{.37$\pm$.03}& \textbf{.39$\pm$.01}& \textbf{.40$\pm$.02}\\ \hline
\end{tabular}}
\caption{Accuracy metrics for the Orientation dataset.}
\label{tab:ori}
\end{table}

In \Cref{sec:corr} we provide a correlation plot across models and prompting strategies, we can observe that the Rhetorics method generate the lowest correlation between strategies.  This evidence points out that the initial bias of the model for inferring change in meaning is very different from the current established methods in linguistic literature.

\section{Discussion}
\label{sec:discussion}
In the boxes below, we present some cherry-picked cases of the reasoning approach with rhetoric devices. This illustrates how the method can improve over no reasoning.

\begin{tcolorbox}[title=Example \#54 cup,boxsep=1pt]
\begin{itemize}
    \setlength\itemsep{0pt}
    \item He drank a \textbf{cup} of coffee.
    \item He borrowed a \textbf{cup} of sugar.
\end{itemize}
In sentence 1, it is used with the sense of a container. 
---
In sentence 2, it is used with the sense of an amount. 
---
"He drank a cup of sugar." Both sentences use 'cup' with different meanings. 
\tcblower
A: different
\end{tcolorbox}

In the example above the \ac{cot} and rhetorics method inference `cup' in the second sentence is used for ``an amount'', instead of ``a container''. However, the rhetorics method differs by generating a zeugma that helps make clear why the senses don't work, as drinking a cup of sugar sounds absurd.

\begin{tcolorbox}[title=Example \#9 battle,boxsep=1pt]
\begin{itemize}
    \setlength\itemsep{0pt}
    \item During the course of this \textbf{battle}, four Canadian soldiers were wounded.
    \item I call on him not to let the \textbf{battle} of hierarchies within the directorates-general prevent this from being implemented.
\end{itemize}
In the first sentence, 'battle' is used literally, referring to a fight or conflict.
---
In the second sentence, 'battle' is used metaphorically to represent a struggle or conflict between different groups or factions.
---
The word 'battle' is related to the first sentence as a metaphor.
\tcblower
A: metaphor
\end{tcolorbox}

For the figurative prompt, the \texttt{Phi-3} model enriched the prompt with a reasonable inference for the second sentence, with this information the model could correctly relate the senses as a metaphorical variation.

\begin{tcolorbox}[title=Example \#167 salt,boxsep=1pt]
\begin{itemize}
    \setlength\itemsep{0pt}
    \item She was \textbf{salty} because of what happened yesterday.
    \item She \textbf{salts} her lectures with jokes.
\end{itemize}
In sentence 1, 'salty' is used to describe someone who is angry or upset.
---
In sentence 2, 'salts' is used to describe the act of adding flavor to something, in this case, a lecture.
---
I'd rather have a lecture with jokes than be angry or upset. The second sense is more positive.
\tcblower
A: positive
\end{tcolorbox}

The orientation prompt forces an order between the senses, making comparing non-trivial perceptions of good and bad feasible. In the example above, the model is instructed to use antanagoge and then impersonate a preference over both scenarios; this allows it to correctly classify the second sense as more positive.

The illustrative examples demonstrate that \acp{llm} can retrieve stored cultural knowledge and enrich prompts, allowing in-context learning to utilize more information for decision-making. However, in some cases, they fail to correctly manipulate senses to produce zeugma, and sometimes the conclusions are inconsistent with the reasoning. We have detailed these failure cases in \Cref{sec:negative}.

This observation aligns with well-documented phenomena: \acp{llm} can hallucinate and generate incorrect reasoning even if they arrive at the correct answer \cite{Ye2022TheUO}. Other failure cases may involve the leakage of evaluation data and the confusion of generalization with memorization, as it is difficult to verify whether an LLM-generated figurative usage explanation is novel or directly derived from training data.

Meaning is a fundamental open question in NLP. While \acp{llm} can often replicate human-like behavior by relying heavily on form, they struggle with simpler tasks that require basic understanding of meaning \cite{Berglund2023TheRC}. Understanding how models deal with meaning in controlled settings (such as comparing a word with itself) is crucial for enabling models to generalize beyond mere form.

While rhetorical devices are standard tools in linguistics, our understanding of their cognitive and psychological effectiveness is still evolving. By using \acp{llm}, we can better test theories and explore how humans relate meanings through language, further advancing our understanding in this area.

\section{Limitations}
Our sentences were carefully selected to enable a concrete understanding of the word's meaning in the context; if the method is applied in under-contextualized sentences, it can result in wrong sense disambiguation, leading to a bad overall performance. We evaluated our approach on LLMs optimized for English; while the methodology applies to other languages, the quality of characterization may differ. LLMs can hallucinate on generating explanations, so the generated rationale may not reflect the correct cognitive likeness between senses even when it correctly characterizes the change.

\section{Conclusion}
\label{sec:conclusion}
In this paper, we exploited the vast amount of cultural knowledge in \acp{llm} to relate and characterize the word's meaning change for different contexts. Hereafter, we experimentally observed that rhetoric devices can help \acp{llm} to mimic human meaning associations, enabling semantic change characterization.

\acp{llm} has many rhetoric examples in its training data, which are not trivially detected. In future works, we expect to explore in depth if \acp{llm} can produce new rhetoric associations or if it's just reproducing its training data, this task can be investigate through non conventionalized metaphor or circumstantial metonymy.

Another important aspect to investigate is how to better prompt language model for figurative reasoning if it can perform better by providing all forms of metaphors (instrumentality, container, etc.) \cite{lakoff2008metaphors} and metonymies (part-whole, cause-effect, etc.) \cite{lakoff2008metaphors} as examples. We also expect to extend the DUREL approach to automatize the characterization of semantic change.


\bibliography{anthology,custom}

\begin{thebibliography}{47}
\providecommand{\natexlab}[1]{#1}

\bibitem[{Agirre and Edmonds(2007)}]{agirre2007word}
Eneko Agirre and Philip Edmonds. 2007.
\newblock \emph{Word sense disambiguation: Algorithms and applications},
  volume~33.
\newblock Springer Science \& Business Media.

\bibitem[{Allan(2013)}]{allan2013oxford}
Keith Allan. 2013.
\newblock \emph{The Oxford handbook of the history of linguistics}.
\newblock OUP Oxford.

\bibitem[{Baccianella et~al.(2010)Baccianella, Esuli, and
  Sebastiani}]{baccianella-etal-2010-sentiwordnet}
Stefano Baccianella, Andrea Esuli, and Fabrizio Sebastiani. 2010.
\newblock \href
  {http://www.lrec-conf.org/proceedings/lrec2010/pdf/769_Paper.pdf}
  {{S}enti{W}ord{N}et 3.0: An enhanced lexical resource for sentiment analysis
  and opinion mining}.
\newblock In \emph{Proceedings of the Seventh International Conference on
  Language Resources and Evaluation ({LREC}'10)}, Valletta, Malta. European
  Language Resources Association (ELRA).

\bibitem[{Berglund et~al.(2023)Berglund, Tong, Kaufmann, Balesni, Stickland,
  Korbak, and Evans}]{Berglund2023TheRC}
Lukas Berglund, Meg Tong, Max Kaufmann, Mikita Balesni, Asa~Cooper Stickland,
  Tomasz Korbak, and Owain Evans. 2023.
\newblock \href {https://api.semanticscholar.org/CorpusID:262083829} {The
  reversal curse: Llms trained on "a is b" fail to learn "b is a"}.
\newblock \emph{ArXiv}, abs/2309.12288.

\bibitem[{Blank(2003)}]{blank2003words}
Andreas Blank. 2003.
\newblock Words and concepts in time: Towards diachronic cognitive
  onomasiology.
\newblock \emph{Trends In Linguistics Studies And Monographs}, 143:37--66.

\bibitem[{Bochkarev et~al.(2022)Bochkarev, Khristoforov, Shevlyakova, and
  Solovyev}]{Bochkarev2022Neural}
Vladimir~V. Bochkarev, Stanislav~V. Khristoforov, Anna~V. Shevlyakova, and
  Valery~D. Solovyev. 2022.
\newblock \href {https://doi.org/10.1109/ACCESS.2022.3186681} {Neural network
  algorithm for detection of new word meanings denoting named entities}.
\newblock \emph{IEEE Access}, 10:68499--68512.

\bibitem[{Camboim~de S{\'a} et~al.(2024)Camboim~de S{\'a}, Anastasiou,
  Da~Silveira, and Pruski}]{camboim-de-sa-etal-2024-socio}
Jader Camboim~de S{\'a}, Dimitra Anastasiou, Marcos Da~Silveira, and C{\'e}dric
  Pruski. 2024.
\newblock \href {https://aclanthology.org/2024.teicai-1.4} {Socio-cultural
  adapted chatbots: Harnessing knowledge graphs and large language models for
  enhanced context awarenes}.
\newblock In \emph{Proceedings of the 1st Worskhop on Towards Ethical and
  Inclusive Conversational AI: Language Attitudes, Linguistic Diversity, and
  Language Rights (TEICAI 2024)}, pages 21--27, St Julians, Malta. Association
  for Computational Linguistics.

\bibitem[{Camboim~de S\'a et~al.(2024)Camboim~de S\'a, Da~Silveira, and
  Pruski}]{Sa2024SurveyIC}
Jader~Martins Camboim~de S\'a, Marcos Da~Silveira, and C{\'e}dric Pruski. 2024.
\newblock \href {https://api.semanticscholar.org/CorpusID:268063910} {Survey in
  characterization of semantic change}.
\newblock \emph{ArXiv}.

\bibitem[{Chen et~al.(2023)Chen, Cheng, Luu, and Bing}]{Chen2023ExploringTP}
Guizhen Chen, Liying Cheng, Anh~Tuan Luu, and Lidong Bing. 2023.
\newblock \href {https://api.semanticscholar.org/CorpusID:265213435} {Exploring
  the potential of large language models in computational argumentation}.
\newblock \emph{ArXiv}, abs/2311.09022.

\bibitem[{Choi et~al.(2021)Choi, Lee, Choi, Park, Lee, Lee, and
  Lee}]{choi-etal-2021-melbert}
Minjin Choi, Sunkyung Lee, Eunseong Choi, Heesoo Park, Junhyuk Lee, Dongwon
  Lee, and Jongwuk Lee. 2021.
\newblock \href {https://doi.org/10.18653/v1/2021.naacl-main.141} {{M}el{BERT}:
  Metaphor detection via contextualized late interaction using metaphorical
  identification theories}.
\newblock In \emph{Proceedings of the 2021 Conference of the North American
  Chapter of the Association for Computational Linguistics: Human Language
  Technologies}, pages 1763--1773, Online. Association for Computational
  Linguistics.

\bibitem[{Dasgupta et~al.(2022)Dasgupta, Lampinen, Chan, Creswell, Kumaran,
  McClelland, and Hill}]{Dasgupta2022LanguageMS}
Ishita Dasgupta, Andrew~Kyle Lampinen, Stephanie C.~Y. Chan, Antonia Creswell,
  Dharshan Kumaran, James~L. McClelland, and Felix Hill. 2022.
\newblock \href {https://api.semanticscholar.org/CorpusID:250526626} {Language
  models show human-like content effects on reasoning}.
\newblock \emph{ArXiv}, abs/2207.07051.

\bibitem[{Fonteyn and Manjavacas(2021)}]{Fonteyn2021AdjustingSA}
Lauren Fonteyn and Enrique Manjavacas. 2021.
\newblock \href {https://api.semanticscholar.org/CorpusID:240290358} {Adjusting
  scope: A computational approach to case-driven research on semantic change}.
\newblock In \emph{Workshop on Computational Humanities Research}.

\bibitem[{Giulianelli et~al.(2023)Giulianelli, Luden, Fernandez, and
  Kutuzov}]{giulianelli-etal-2023-interpretable}
Mario Giulianelli, Iris Luden, Raquel Fernandez, and Andrey Kutuzov. 2023.
\newblock \href {https://doi.org/10.18653/v1/2023.acl-long.176} {Interpretable
  word sense representations via definition generation: The case of semantic
  change analysis}.
\newblock In \emph{Proceedings of the 61st Annual Meeting of the Association
  for Computational Linguistics (Volume 1: Long Papers)}, pages 3130--3148,
  Toronto, Canada. Association for Computational Linguistics.

\bibitem[{Hengchen et~al.(2021)Hengchen, Tahmasebi, Schlechtweg, and
  Dubossarsky}]{Hengchen2021ChallengesFC}
Simon Hengchen, Nina Tahmasebi, Dominik Schlechtweg, and Haim Dubossarsky.
  2021.
\newblock Challenges for computational lexical semantic change.
\newblock \emph{ArXiv}, abs/2101.07668.

\bibitem[{Huang and Chang(2023)}]{huang-chang-2023-towards}
Jie Huang and Kevin Chen-Chuan Chang. 2023.
\newblock \href {https://doi.org/10.18653/v1/2023.findings-acl.67} {Towards
  reasoning in large language models: A survey}.
\newblock In \emph{Findings of the Association for Computational Linguistics:
  ACL 2023}, pages 1049--1065, Toronto, Canada. Association for Computational
  Linguistics.

\bibitem[{Juvonen and Koptjevskaja-Tamm(2016)}]{juvonen2016lexical}
P{\"a}ivi Juvonen and Maria Koptjevskaja-Tamm. 2016.
\newblock \emph{The lexical typology of semantic shifts}, volume~58.
\newblock Walter de Gruyter GmbH \& Co KG.

\bibitem[{Kavumba et~al.(2023)Kavumba, Brassard, Heinzerling, and
  Inui}]{kavumba-etal-2023-prompting}
Pride Kavumba, Ana Brassard, Benjamin Heinzerling, and Kentaro Inui. 2023.
\newblock \href {https://doi.org/10.18653/v1/2023.findings-eacl.162} {Prompting
  for explanations improves adversarial {NLI}. is this true? {Yes} it is {true}
  because {it weakens superficial cues}}.
\newblock In \emph{Findings of the Association for Computational Linguistics:
  EACL 2023}, pages 2165--2180, Dubrovnik, Croatia. Association for
  Computational Linguistics.

\bibitem[{Kearns(2006)}]{kearns2006lexical}
Kate Kearns. 2006.
\newblock \emph{The Handbook of English Linguistics}.
\newblock Wiley Online Library.

\bibitem[{Kilgarriff(1997)}]{kilgarriff1997don}
Adam Kilgarriff. 1997.
\newblock I don’t believe in word senses.
\newblock \emph{Computers and the Humanities}, 31:91--113.

\bibitem[{Lakoff and Johnson(2008)}]{lakoff2008metaphors}
George Lakoff and Mark Johnson. 2008.
\newblock \emph{Metaphors we live by}.
\newblock University of Chicago press.

\bibitem[{Liu et~al.(2022)Liu, Cui, Zheng, and Neubig}]{liu-etal-2022-testing}
Emmy Liu, Chenxuan Cui, Kenneth Zheng, and Graham Neubig. 2022.
\newblock \href {https://doi.org/10.18653/v1/2022.naacl-main.330} {Testing the
  ability of language models to interpret figurative language}.
\newblock In \emph{Proceedings of the 2022 Conference of the North American
  Chapter of the Association for Computational Linguistics: Human Language
  Technologies}, pages 4437--4452, Seattle, United States. Association for
  Computational Linguistics.

\bibitem[{Lyu et~al.(2022)Lyu, Zhou, and Ji}]{lyu-etal-2022-mllabs}
Chenyang Lyu, Yongxin Zhou, and Tianbo Ji. 2022.
\newblock \href {https://doi.org/10.18653/v1/2022.evonlp-1.1} {{MLL}abs-{LIG}
  at {T}empo{W}i{C} 2022: A generative approach for examining temporal meaning
  shift}.
\newblock In \emph{Proceedings of the First Workshop on Ever Evolving NLP
  (EvoNLP)}, pages 1--6, Abu Dhabi, United Arab Emirates (Hybrid). Association
  for Computational Linguistics.

\bibitem[{Maudslay and Teufel(2022)}]{maudslay-teufel-2022-metaphorical}
Rowan~Hall Maudslay and Simone Teufel. 2022.
\newblock \href {https://aclanthology.org/2022.coling-1.7} {Metaphorical
  polysemy detection: Conventional metaphor meets word sense disambiguation}.
\newblock In \emph{Proceedings of the 29th International Conference on
  Computational Linguistics}, pages 65--77, Gyeongju, Republic of Korea.
  International Committee on Computational Linguistics.

\bibitem[{Miller(1995)}]{Miller1995WordNetAL}
George~A. Miller. 1995.
\newblock \href {https://api.semanticscholar.org/CorpusID:1671874} {Wordnet: A
  lexical database for english}.
\newblock \emph{Commun. ACM}, 38:39--41.

\bibitem[{Mitra et~al.(2023)Mitra, Corro, Mahajan, Codas, Simoes, Agrawal,
  Chen, Razdaibiedina, Jones, Aggarwal, Palangi, Zheng, Rosset, Khanpour, and
  Awadallah}]{Mitra2023Orca2T}
Arindam Mitra, Luciano~Del Corro, Shweti Mahajan, Andres Codas, Clarisse
  Simoes, Sahaj Agrawal, Xuxi Chen, Anastasia Razdaibiedina, Erik Jones, Kriti
  Aggarwal, Hamid Palangi, Guoqing Zheng, Corby Rosset, Hamed Khanpour, and
  Ahmed Awadallah. 2023.
\newblock \href {https://api.semanticscholar.org/CorpusID:265295592} {Orca 2:
  Teaching small language models how to reason}.
\newblock \emph{ArXiv}, abs/2311.11045.

\bibitem[{OpenAI(2023)}]{Achiam2023GPT4TR}
OpenAI. 2023.
\newblock \href {https://api.semanticscholar.org/CorpusID:257532815} {Gpt-4
  technical report}.
\newblock \emph{ArXiv}.

\bibitem[{Periti et~al.(2024)Periti, Cassotti, Dubossarsky, and
  Tahmasebi}]{Periti2024AnalyzingSC}
Francesco Periti, Pierluigi Cassotti, Haim Dubossarsky, and Nina Tahmasebi.
  2024.
\newblock \href {https://api.semanticscholar.org/CorpusID:269448622} {Analyzing
  semantic change through lexical replacements}.
\newblock \emph{ArXiv}, abs/2404.18570.

\bibitem[{Petroni et~al.(2019)Petroni, Rockt{\"a}schel, Riedel, Lewis, Bakhtin,
  Wu, and Miller}]{petroni-etal-2019-language}
Fabio Petroni, Tim Rockt{\"a}schel, Sebastian Riedel, Patrick Lewis, Anton
  Bakhtin, Yuxiang Wu, and Alexander Miller. 2019.
\newblock \href {https://doi.org/10.18653/v1/D19-1250} {Language models as
  knowledge bases?}
\newblock In \emph{Proceedings of the 2019 Conference on Empirical Methods in
  Natural Language Processing and the 9th International Joint Conference on
  Natural Language Processing (EMNLP-IJCNLP)}, pages 2463--2473, Hong Kong,
  China. Association for Computational Linguistics.

\bibitem[{Pilehvar and
  Camacho-Collados(2019)}]{pilehvar-camacho-collados-2019-wic}
Mohammad~Taher Pilehvar and Jose Camacho-Collados. 2019.
\newblock \href {https://doi.org/10.18653/v1/N19-1128} {{W}i{C}: the
  word-in-context dataset for evaluating context-sensitive meaning
  representations}.
\newblock In \emph{Proceedings of the 2019 Conference of the North {A}merican
  Chapter of the Association for Computational Linguistics: Human Language
  Technologies, Volume 1 (Long and Short Papers)}, pages 1267--1273,
  Minneapolis, Minnesota. Association for Computational Linguistics.

\bibitem[{Pinker(2003)}]{pinker2003language}
Steven Pinker. 2003.
\newblock \emph{The language instinct: How the mind creates language}.
\newblock Penguin UK.

\bibitem[{Pinker and Bloom(1990)}]{Pinker1990NaturalLA}
Steven Pinker and Paula~Jorde Bloom. 1990.
\newblock Natural language and natural selection.
\newblock \emph{Behavioral and Brain Sciences}, 13:707 -- 727.

\bibitem[{Pivovarova and Kutuzov(2021)}]{Pivovarova2021RuShiftEvalAS}
Lidia Pivovarova and Andrey Kutuzov. 2021.
\newblock \href {https://api.semanticscholar.org/CorpusID:240007326}
  {Rushifteval: a shared task on semantic shift detection for russian}.
\newblock \emph{ArXiv}.

\bibitem[{Raganato et~al.(2017)Raganato, Camacho-Collados, and
  Navigli}]{raganato-etal-2017-word}
Alessandro Raganato, Jose Camacho-Collados, and Roberto Navigli. 2017.
\newblock \href {https://aclanthology.org/E17-1010} {Word sense disambiguation:
  A unified evaluation framework and empirical comparison}.
\newblock In \emph{Proceedings of the 15th Conference of the {E}uropean Chapter
  of the Association for Computational Linguistics: Volume 1, Long Papers},
  pages 99--110, Valencia, Spain. Association for Computational Linguistics.

\bibitem[{Schlechtweg et~al.(2020)Schlechtweg, McGillivray, Hengchen,
  Dubossarsky, and Tahmasebi}]{schlechtweg-etal-2020-semeval}
Dominik Schlechtweg, Barbara McGillivray, Simon Hengchen, Haim Dubossarsky, and
  Nina Tahmasebi. 2020.
\newblock \href {https://doi.org/10.18653/v1/2020.semeval-1.1}
  {{S}em{E}val-2020 task 1: Unsupervised lexical semantic change detection}.
\newblock In \emph{Proceedings of the Fourteenth Workshop on Semantic
  Evaluation}, pages 1--23, Barcelona (online). International Committee for
  Computational Linguistics.

\bibitem[{Schlechtweg et~al.(2024)Schlechtweg, Virk, Sander, Sk{\"o}ldberg,
  Theuer~Linke, Zhang, Tahmasebi, Kuhn, and Schulte
  Im~Walde}]{schlechtweg-etal-2024-durel}
Dominik Schlechtweg, Shafqat~Mumtaz Virk, Pauline Sander, Emma Sk{\"o}ldberg,
  Lukas Theuer~Linke, Tuo Zhang, Nina Tahmasebi, Jonas Kuhn, and Sabine Schulte
  Im~Walde. 2024.
\newblock \href {https://aclanthology.org/2024.eacl-demo.15} {The {DUR}el
  annotation tool: Human and computational measurement of semantic proximity,
  sense clusters and semantic change}.
\newblock In \emph{Proceedings of the 18th Conference of the European Chapter
  of the Association for Computational Linguistics: System Demonstrations},
  pages 137--149, St. Julians, Malta. Association for Computational
  Linguistics.

\bibitem[{Steen et~al.(2007)Steen, Cameron, Cienki, Crisp, Deignan, Gibbs,
  Grady, K{\"o}vecses, Low, and Semino}]{Steen2007MIPAM}
Gerard Steen, Lynne Cameron, Alan Cienki, Peter Crisp, Alice Deignan,
  Raymond~W. Gibbs, Joe Grady, Zolt{\'a}n K{\"o}vecses, Graham~David Low, and
  Elena Semino. 2007.
\newblock \href {https://api.semanticscholar.org/CorpusID:142786072} {Mip: A
  method for identifying metaphorically used words in discourse}.
\newblock \emph{Metaphor and Symbol}, 22:1--39.

\bibitem[{Strachan et~al.(2024)Strachan, Albergo, Borghini, Pansardi, Scaliti,
  Gupta, Saxena, Rufo, Panzeri, Manzi, Graziano, and
  Becchio}]{Strachan2024TestingTO}
James W.~A. Strachan, Dalila Albergo, Giulia Borghini, Oriana Pansardi, Eugenio
  Scaliti, Saurabh Gupta, Krati Saxena, Alessandro Rufo, Stefano Panzeri, Guido
  Manzi, Michael S~A Graziano, and Cristina Becchio. 2024.
\newblock \href {https://api.semanticscholar.org/CorpusID:269928651} {Testing
  theory of mind in large language models and humans.}
\newblock \emph{Nature human behaviour}.

\bibitem[{Tahmasebi et~al.(2018)Tahmasebi, Borin, and
  Jatowt}]{Tahmasebi2018SurveyOC}
Nina Tahmasebi, Lars Borin, and Adam Jatowt. 2018.
\newblock Survey of computational approaches to lexical semantic change.
\newblock \emph{arXiv: Computation and Language}.

\bibitem[{Tang et~al.(2013)Tang, Qu, and Chen}]{Tang2013SemanticCC}
Xuri Tang, Weiguang Qu, and Xiaohe Chen. 2013.
\newblock Semantic change computation: A successive approach.
\newblock \emph{World Wide Web}, 19:375--415.

\bibitem[{Traugott(2017)}]{traugott2017semantic}
Elizabeth~Closs Traugott. 2017.
\newblock \href {https://doi.org/10.1093/acrefore/9780199384655.013.323}
  {Semantic change}.

\bibitem[{Wang and Choi(2023)}]{Wang2023LargeLM}
Ruiyu Wang and Matthew Choi. 2023.
\newblock \href {https://api.semanticscholar.org/CorpusID:266163095} {Large
  language models on lexical semantic change detection: An evaluation}.
\newblock \emph{ArXiv}, abs/2312.06002.

\bibitem[{Webb et~al.(2022)Webb, Holyoak, and Lu}]{Webb2022EmergentAR}
Taylor~W. Webb, Keith~J. Holyoak, and Hongjing Lu. 2022.
\newblock \href {https://api.semanticscholar.org/CorpusID:254854575} {Emergent
  analogical reasoning in large language models}.
\newblock \emph{Nature Human Behaviour}, 7:1526 -- 1541.

\bibitem[{Wei et~al.(2022)Wei, Wang, Schuurmans, Bosma, Xia, Chi, Le, Zhou
  et~al.}]{Wei2022ChainOT}
Jason Wei, Xuezhi Wang, Dale Schuurmans, Maarten Bosma, Fei Xia, Ed~Chi, Quoc~V
  Le, Denny Zhou, et~al. 2022.
\newblock Chain-of-thought prompting elicits reasoning in large language
  models.
\newblock \emph{Advances in neural information processing systems},
  35:24824--24837.

\bibitem[{Wiebe and Mihalcea(2006)}]{wiebe-mihalcea-2006-word}
Janyce Wiebe and Rada Mihalcea. 2006.
\newblock \href {https://doi.org/10.3115/1220175.1220309} {Word sense and
  subjectivity}.
\newblock In \emph{Proceedings of the 21st International Conference on
  Computational Linguistics and 44th Annual Meeting of the Association for
  Computational Linguistics}, pages 1065--1072, Sydney, Australia. Association
  for Computational Linguistics.

\bibitem[{Ye and Durrett(2022)}]{Ye2022TheUO}
Xi~Ye and Greg Durrett. 2022.
\newblock \href {https://api.semanticscholar.org/CorpusID:252873674} {The
  unreliability of explanations in few-shot prompting for textual reasoning}.
\newblock In \emph{Neural Information Processing Systems}.

\bibitem[{Zamora-Reina et~al.(2022)Zamora-Reina, Bravo-Marquez, and
  Schlechtweg}]{ZamoraReina2022blackLSCDiscoveryST}
Frank~D. Zamora-Reina, Felipe Bravo-Marquez, and Dominik Schlechtweg. 2022.
\newblock \href {https://api.semanticscholar.org/CorpusID:248780554}
  {Lscdiscovery: A shared task on semantic change discovery and detection in
  spanish}.
\newblock \emph{ArXiv}, abs/2205.06691.

\bibitem[{Zhang et~al.(2023)Zhang, Mao, He, and
  Cambria}]{Zhang2023NeuroSymbolicSA}
Xulang Zhang, Rui Mao, Kai He, and Erik Cambria. 2023.
\newblock \href {https://api.semanticscholar.org/CorpusID:264486324}
  {Neuro-symbolic sentiment analysis with dynamic word sense disambiguation}.
\newblock In \emph{Conference on Empirical Methods in Natural Language
  Processing}.

\end{thebibliography}

\clearpage
\appendix

\section{Negative Examples}
\label{sec:negative}
In this section, we present some examples where the model failed to produce the correct form or to give an answer over the correctly produced reasoning.
\subsection{Dimension}
Below are some examples of dimension data:

\begin{tcolorbox}[title=Example \#44 come,fontupper=\small]
\begin{itemize}
    \item \textbf{Come} out of the closet!
    \item He \textbf{came} singing down the road.
\end{itemize}
In sentence 1 it is used with the sense of emerge. 
---
In sentence 2 it is used with the sense of move in a certain direction.
---
"He came out of the closet singing." The word 'come' doesn't make a bad pun, but it has different senses.  It is still different.
\tcblower
A: different
\end{tcolorbox}
In the example above, the zeugma was not correctly constructed, and it ignored the instruction, which only makes sense if it makes a bad pun.

\begin{tcolorbox}[title=Example \#67 do,fontupper=\small]
\begin{itemize}
    \item All you ever \textbf{do} is surf the Internet.
    \item A  'B' grade \textbf{doesn't} suffice to get me into medical school.
\end{itemize}
In sentence 1 it is used with the sense of an action. 
---
In sentence 2 is used with the sense of to suffice, to be enough. 
---
"A' B' grade does not suffice for me to do what I want." The word 'do' makes a bad pun, so is different. 
\tcblower
A: different
\end{tcolorbox}
In the example above, it failed to build a zeugma using both contexts, even though the answer is correct.

\subsection{Orientation}
Below are some examples of orientation data.

\begin{tcolorbox}[title=Example \#67 do,fontupper=\small]
\begin{itemize}
    \item He do not resent they supervision or person 's sometimes tiring \textbf{sympathy}.
    \item The group exhibit more concern over the possibility of the abolitionist make a martyr of person than it do over the development of \textbf{sympathy} for he.
\end{itemize}
In the first sentence, 'sympathy' is used in a negative context, indicating annoyance or irritation.
---
In the second sentence, 'sympathy' is used to express understanding or concern.
---
Expressing understanding or concern (second sentence) is more positive than feeling annoyed or irritated. 
\tcblower
A: positive
\end{tcolorbox}
While the answer was correct, in the example above, the model didn't use antithesis for ordering senses.

\section{Inference Settings}
Given hardware constraints, we used the gguf version of the models with \texttt{llama.cpp}\footnote{\url{https://github.com/ggerganov/llama.cpp}} library. In the \texttt{Phi-3} (mini version), we used the 4k context with fp16 quantization. \texttt{LLaMA-3-8b} we used 8bit quantization and \texttt{LLaMA-3-70b}, 2bit quantization. All model weights were obtained from HuggingFace\footnote{\url{https://huggingface.co/models}}.

We did all the experiments on a Tesla V100 with 32GB RAM. The inference for all the models and prompts took less than four days.

\section{Correlation Across Prompt Strategies and Models}
\label{sec:corr}
In this section, we present the correlation plot between judgments across different models and different prompts.

\begin{figure}[ht]
    \centering
    \includegraphics[width=\columnwidth]{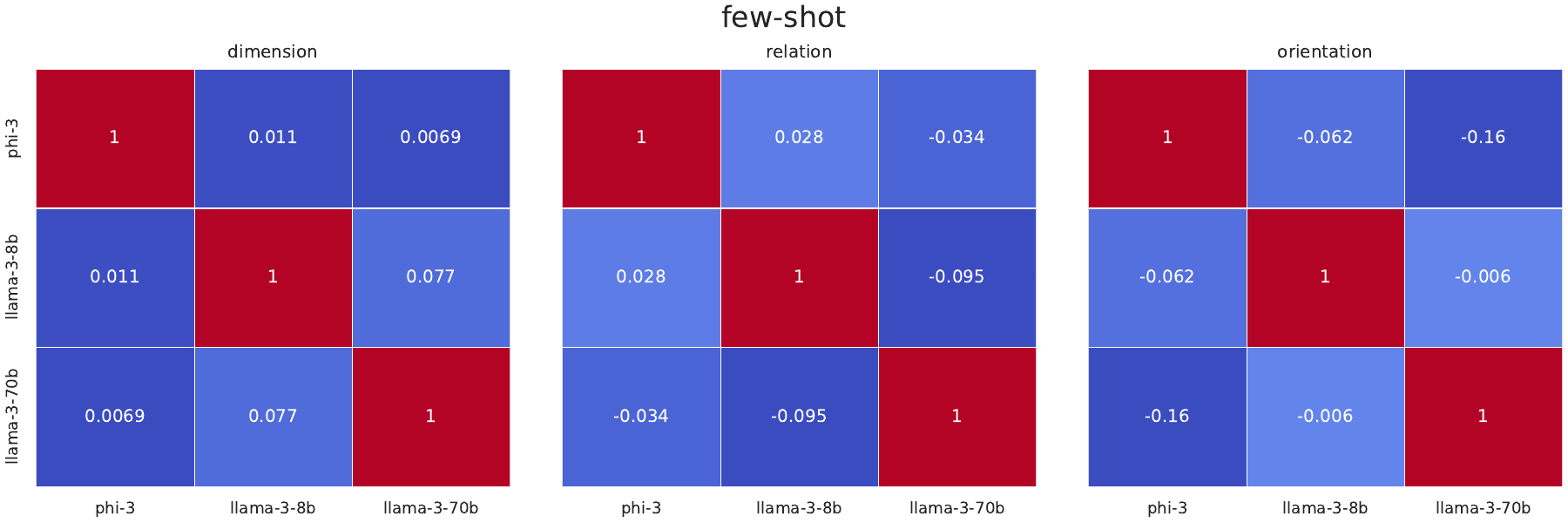}
    \caption{Correlation for Few-shot prompting.}
    \label{fig:corr-few}
\end{figure}

\begin{figure}[ht]
    \centering
    \includegraphics[width=\columnwidth]{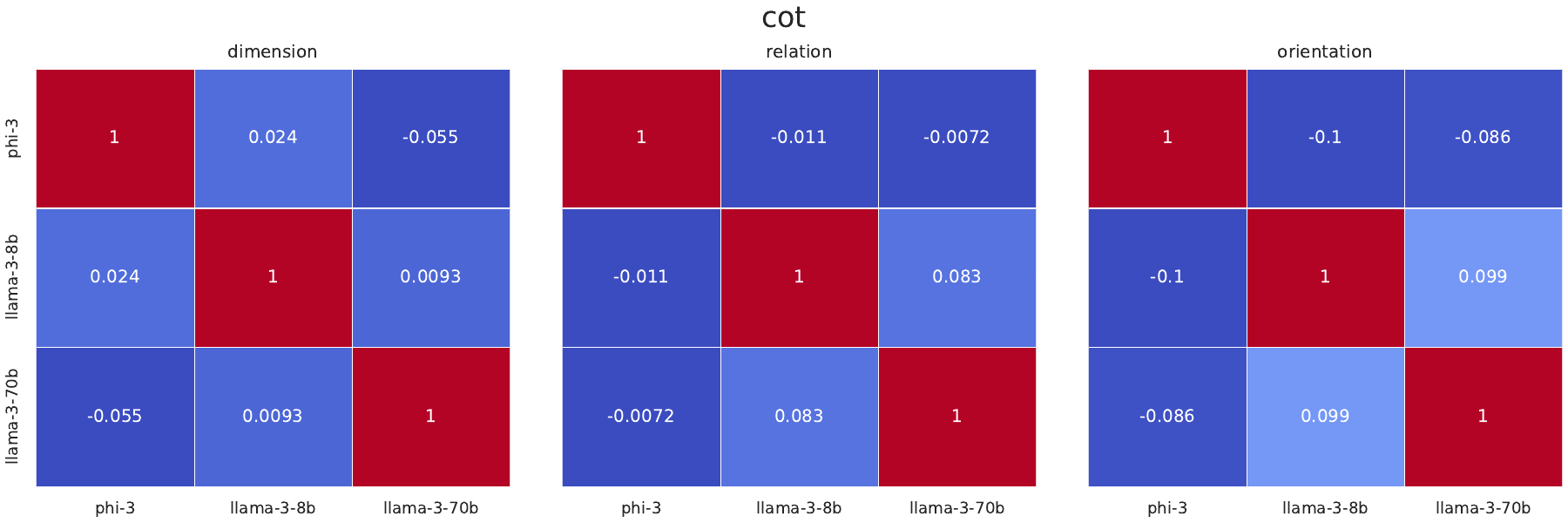}
    \caption{Correlation for CoT prompting across models.}
    \label{fig:corr-cot}
\end{figure}

\begin{figure}[ht]
    \centering
    \includegraphics[width=\columnwidth]{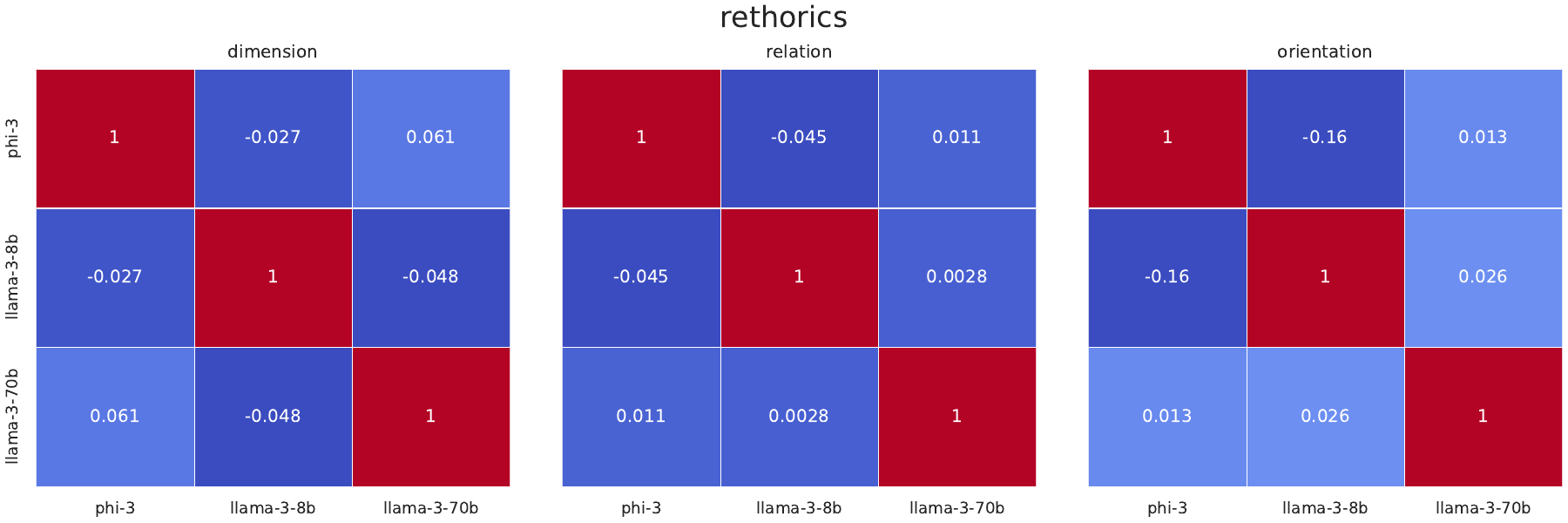}
    \caption{Correlation for Rhetorics prompting across models.}
    \label{fig:corr-ret}
\end{figure}

\begin{figure}[ht]
    \centering
    \includegraphics[width=\columnwidth]{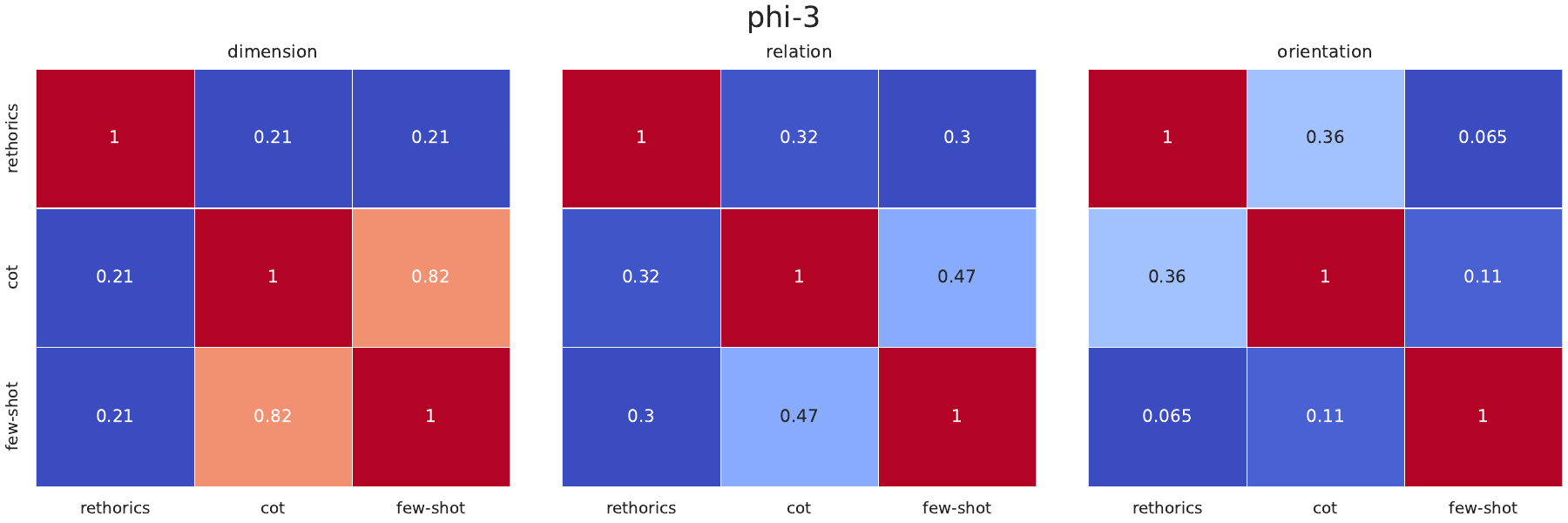}
    \caption{Correlation for Phi-3 model across strategies.}
    \label{fig:corr-phi}
\end{figure}

\begin{figure}[ht]
    \centering
    \includegraphics[width=\columnwidth]{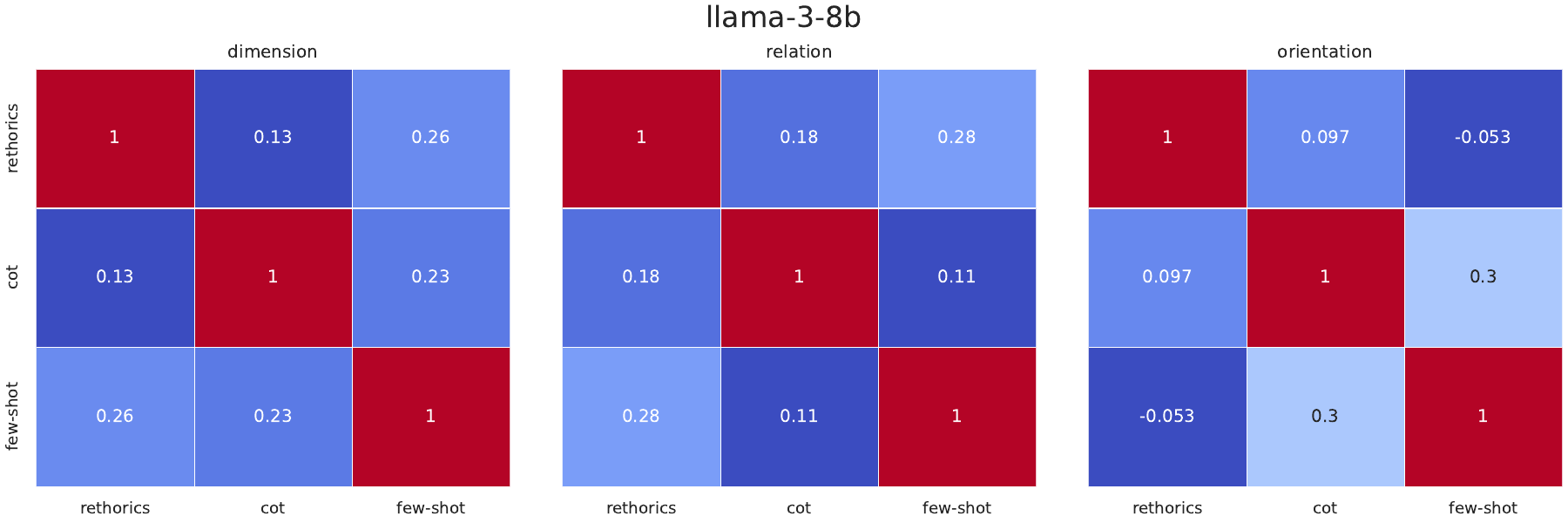}
    \caption{Correlation for LlaMA-3-8b model across strategies.}
    \label{fig:corr-8b}
\end{figure}

\begin{figure}[ht]
    \centering
    \includegraphics[width=\columnwidth]{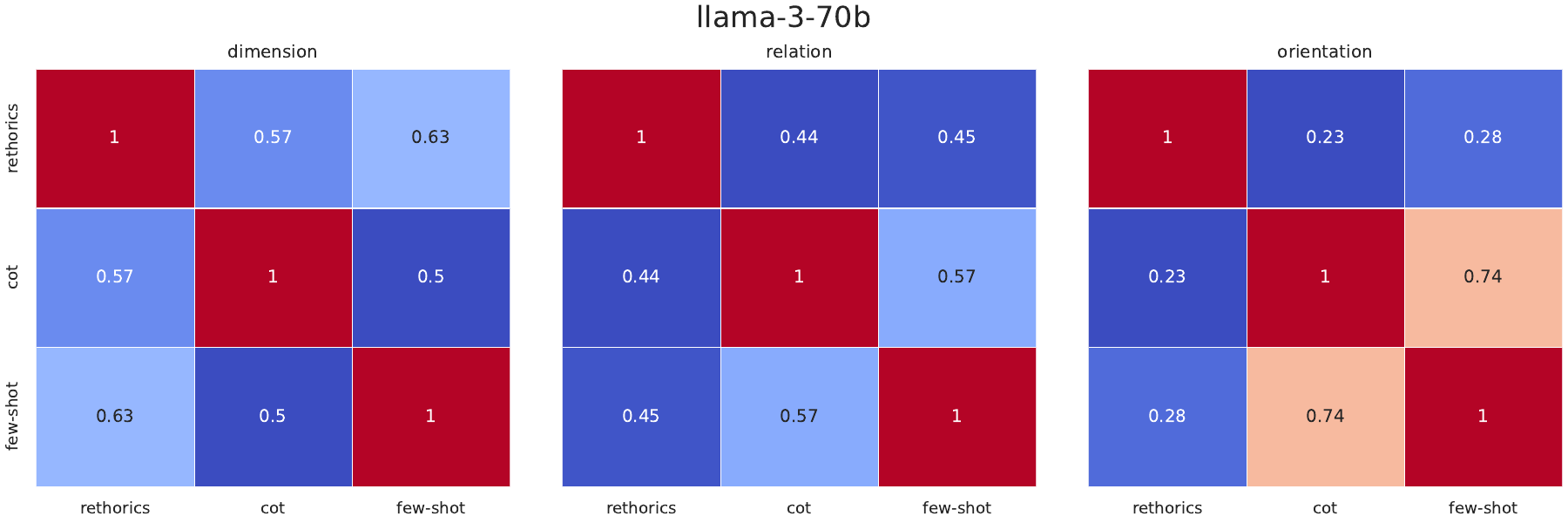}
    \caption{Correlation for LlaMA-3-70b model across strategies.}
    \label{fig:corr-70b}
\end{figure}

In \Cref{fig:corr-70b} we observe that few-shot and CoT approaches are highly correlated.

\section{Detailed Prompts}
\label{sec:prompt}
In the figures below, we show the detailed prompt for each type of characterization.
\begin{figure}[ht]
    \centering
    \begin{tcolorbox}[title=Sense Differentiation, fontupper=\small]
    You are presented with two sentences that both contain a specific word. Your task is to analyze how this word is used in each sentence and determine if its usage in the second sentence represents the same sense with respect to its use in the first sentence. Follow these steps to complete the task:
    \begin{itemize}
    \item Step 1. Describe the meaning of the word in the first sentence.
    \item Step 2. Describe the meaning of the word in the second sentence.
    \item Step 3. Write a sentence that joins both sentences using zeugma and the same shared word while preserving the same sense. If the construction make a bad pun, the words have different sense.
    \item Step 4. Based on the previous reasoning give your final answer: 'identical' or 'different.'
    \end{itemize}
    \tcblower
    \textcolor{gray}{[Few-shot examples.]}
    \end{tcolorbox}
    \caption{Prompt for sense differentiation in the dimension dataset. The model is instructed to perform a zeugma association between senses to reason if it has a identical or different sense}
    \label{fig:dim_prompt}
\end{figure}

\begin{figure}[ht]
    \centering
    \begin{tcolorbox}[title=Sense Figurativeness,fontupper=\small]
    You are presented with two sentences that both contain a specific word. Your task is to analyze how this word is used in each sentence and determine if its usage in the second sentence represents a metaphor or a metonymy with respect to its use in the first sentence. Follow these steps to complete the task:
    \begin{itemize}
    \item Step 1. Describe the meaning of the word in the first sentence.
    \item Step 2. Describe the meaning of the word in the second sentence.
    \item Step 3. Compare the uses, determining if the second is related as a metaphor (where the word is used in a similar but non-literal sense), as a metonymy (where the word represents something closely related to or associated with it), or unrelated, used with a different sense.
    \item Step 4. Based on the third reasoning, write the final answer, 'metaphor', 'metonymy', or 'unrelated.'
    \end{itemize}
    \tcblower
    \textcolor{gray}{[Few-shot examples.]}
    \end{tcolorbox}
    \caption{Prompt for figurative sense in the relation dataset. The model is instructed to relate the meanings by association or similarity.}
    \label{fig:rel_prompt}
\end{figure}

\begin{figure}[ht]
    \centering
    \begin{tcolorbox}[title=Sense Orientation,fontupper=\small]
    You will be provided with two sentences that share a common word used with different senses. Your task is to describe if the second sense for the word is more positive then the first. Follow this instructions to execute the task:
    \begin{itemize}
    \item Step 1. Describe the meaning of the word in the first sentence.
    \item Step 2. Describe the meaning of the word in the second sentence.
    \item Step 3. Leverage the rhetorical strategy of antithesis, contrasting a negative with a positive, to weigh why one meaning might be more favorable than the other, or if they stand neutral.
    \item Step 4. Based on the third reasoning, write the final answer 'negative', 'positive', or 'neutral.'
    \end{itemize}
    \tcblower
    \textcolor{gray}{[Few-shot examples.]}
    \end{tcolorbox}
    \caption{Prompt for sense orientation in the orientation dataset. By using antithesis the model should order the senses polarity using `personal preference' argumentation.}
    \label{fig:ori_prompt}
\end{figure}

\section{Annotation}
To obtain sentiment labels for the orientation data we relied on human annotation from volunteer students from a University in Europe where students have different backgrounds and different native language but English is used as the main language in their studies. The annotations were anonymously collected.

We first provided the annotators with the agreement terms: ``This is a study on sentiment perception of polysemous words. This data will be freely available for research purposes.
Inside you'll be asked to rate how the feeling varies for a word in different sentences.
Your answers will be completely anonymous. COMPANY will not collect your personal data through this questionnaire and will not be able to identify you based on your answers. For more information about COMPANY’s privacy notice please visit our webpage at: URL''

\begin{figure}
    \centering
    \includegraphics[width=\columnwidth]{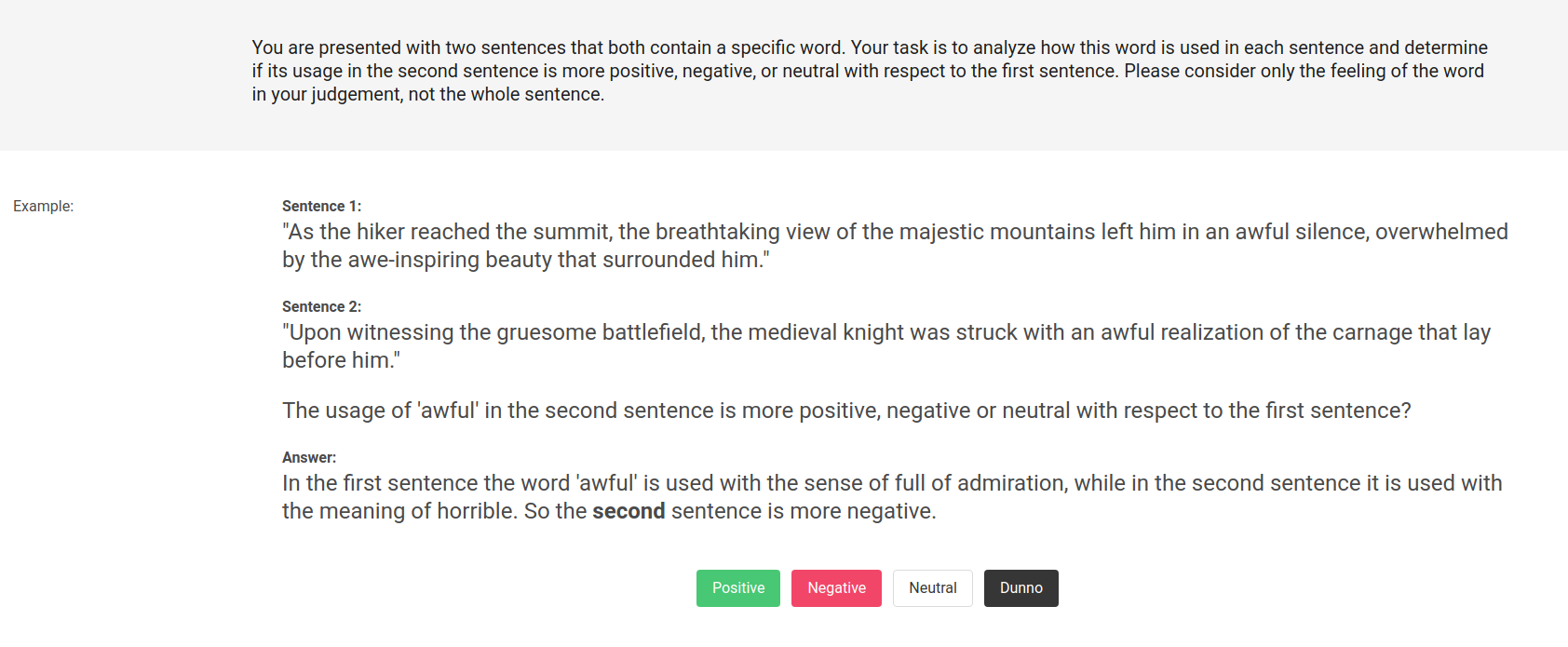}
    \caption{Training screen}
    \label{fig:trainingscreen}
\end{figure}

Then we presented a training screen in
\Cref{fig:trainingscreen}. 

\begin{figure}
    \centering
    \includegraphics[width=\columnwidth]{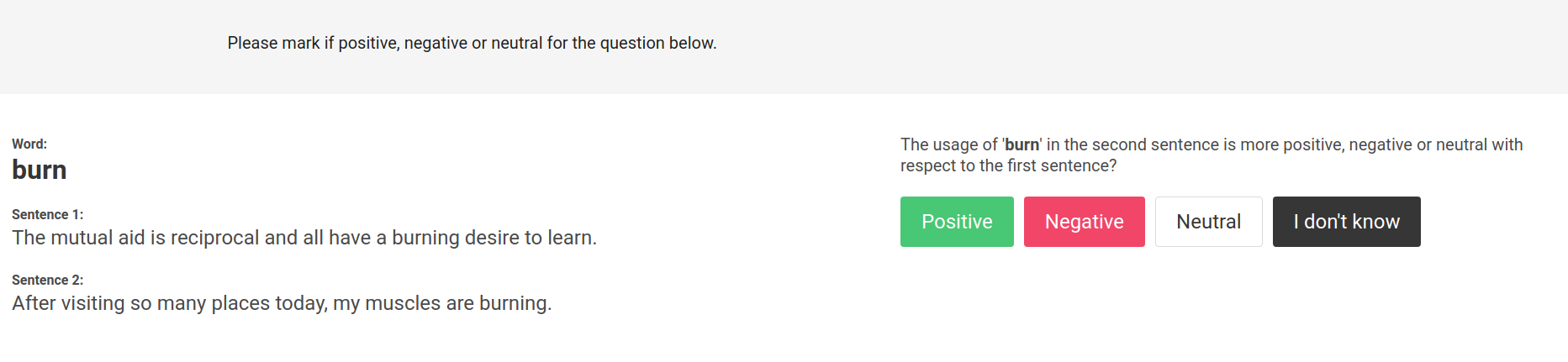}
    \caption{Annotation screen for sentiments.}
    \label{fig:annotation}
\end{figure}

We prompted the annotators for sentiment analysis with screen \Cref{fig:annotation}.

\section{Ai Assistants In Research Or Writing}
As our native language is not English, we used AI assistants like Grammarly, ChatGPT, and Gemini to improve vocabulary, grammar, and readability of this documents and the prompts. We also checked all generated text for inconsistencies with the original intent and fixed them properly when identified.

\end{document}